%% file: main.tex
\documentclass{article}
\usepackage{fullpage}

\input packages.tex
\input macros.tex

\input{math_notation_for_this_paper}

\usepackage{natbib}
\setcitestyle{authoryear,open={(},close={)}} %Citation-related commands

\usepackage{xcolor}
\usepackage{listings}
\makeatletter
\def\adl@drawiv#1#2#3{%
        \hskip.5\tabcolsep
        \xleaders#3{#2.5\@tempdimb #1{1}#2.5\@tempdimb}%
                #2\z@ plus1fil minus1fil\relax
        \hskip.5\tabcolsep}
\newcommand{\cdashlinelr}[1]{%
  \noalign{\vskip\aboverulesep
           \global\let\@dashdrawstore\adl@draw
           \global\let\adl@draw\adl@drawiv}
  \cdashline{#1}
  \noalign{\global\let\adl@draw\@dashdrawstore
           \vskip\belowrulesep}}
\makeatother

\definecolor{upforestgreen}{rgb}{0.6, 0.8, 0.2}

\usepackage{chngcntr}
\usepackage{adjustbox}

\usepackage{wrapfig}
\usepackage{arydshln}

\usepackage{tcolorbox}

\newcommand*\samethanks[1][\value{footnote}]{\footnotemark[#1]}

\begin{document}

\title{Random-LTD: Random and Layerwise Token Dropping Brings Efficient Training for Large-scale Transformers}

\author{
Zhewei Yao\thanks{Equal contribution. Code will be released soon as a part of \url{https://github.com/microsoft/DeepSpeed}}~, Xiaoxia Wu\samethanks~, Conglong Li, Connor Holmes\\
Minjia Zhang, Cheng Li, Yuxiong He \\
Microsoft \\ 
{\tt \small\{zheweiyao,  xiaoxiawu, conglong.li, connorholmes, minjiaz, chengli1, yuxhe\}@microsoft.com}
}

\maketitle

%%%%%%%% BODY TEXT
\input{_s0_abstract.tex}
\input{_s1_intro.tex}

\input{_s2_related_work.tex}
\input{_s3_method.tex}

\input{_s4_results.tex}

\input{_s5_discussion.tex}

\input{_s6_conclusion.tex}

\clearpage
{
\bibliography{ref.bib}
}

\clearpage
\onecolumn
\input _s7_appendix.tex

\end{document}

%% file: packages.tex
\usepackage[utf8]{inputenc} % allow utf-8 input
\usepackage[T1]{fontenc}    % use 8-bit T1 fonts
\usepackage{url}            % simple URL typesetting
\usepackage{booktabs}       % professional-quality tables
\usepackage{amsfonts}       % blackboard math symbols
\usepackage{nicefrac}       % compact symbols for 1/2, etc.
\usepackage{microtype}      % microtypography
\usepackage{soul}
\usepackage{graphicx}
\usepackage{amsmath}
\usepackage{outlines}
\usepackage{xurl}

\usepackage[colorlinks=true,
citecolor=blue,
filecolor=black,
linkcolor=blue,
urlcolor=blue]{hyperref}

\input{math_commands.tex}

\usepackage{multirow}
\usepackage{paralist}

\usepackage{multicol}
\usepackage{diagbox}

\usepackage[ruled,noend]{algorithm2e}

\SetCommentSty{mycommfont}

\usepackage{here}

\usepackage{amsmath,amssymb,amsfonts,amsbsy,amsfonts,latexsym}
\usepackage{makecell}
\usepackage{xcolor}
\usepackage{colortbl}

\usepackage{tabularx,colortbl,xcolor}
\usepackage[normalem]{ulem}
\useunder{\uline}{\ul}{}

\usepackage{enumitem}

\usepackage{xparse}% http://ctan.org/pkg/xparse

\SetKwInput{KwInput}{Input}
\SetKwInput{KwRequire}{Require}

\usepackage{longtable}

%% file: math_commands.tex
\usepackage{amsmath,amsfonts,bm}

\def\eqref#1{equation~\ref{#1}}
\def\1{\bm{1}}

\DeclareMathAlphabet{\mathsfit}{\encodingdefault}{\sfdefault}{m}{sl}
\SetMathAlphabet{\mathsfit}{bold}{\encodingdefault}{\sfdefault}{bx}{n}

%% file: macros.tex
\NewDocumentCommand{\var}{O{s} m O{}}{%
  \ensuremath{#1_{#2}^{#3}}% add \vphantom{<bizarre sup>}
}
\usepackage{siunitx}

\newcommand{\commentout}[1]{}

\definecolor{light-gray}{gray}{0.80}

\newcommand\appref{Appendix~\ref}

\newcommand\fref{Fig.~\ref}
\newcommand\tref{Tab.~\ref}
\newcommand\sref{Section~\ref}

\usepackage{amsthm}

\usepackage{amsthm}

\newtheorem{mytheorem}{Theorem}
\newtheorem{assumption}[mytheorem]{Assumption}
\newtheorem{lemma}[mytheorem]{Lemma}
\newtheorem{remk}[mytheorem]{Remark}

%% file: math_notation_for_this_paper.tex
\newcommand{\OURS}{random-LTD\xspace}
\newcommand{\tokenbypass}{TokenBypass\xspace}
\newcommand{\ltd}{LTD\xspace}

\newcommand{\mha}{MHA\xspace}
\newcommand{\ffc}{FFC\xspace}
\newcommand{\bert}{BERT\xspace}
\newcommand{\vit}{ViT\xspace}
\newcommand{\gpt}{GPT\xspace}
\newcommand{\pslg}{MSLG\xspace}

\newcommand{\layertoken}{LayerToken\xspace}
\newcommand{\layertokens}{LayerTokens\xspace}
\newcommand{\layertokenlr}{LayerToken LR\xspace}
\def\bertbase{BERT$_{\text{base}}$\xspace}
\def\bertlarge{BERT$_{\text{large}}$\xspace}
\newcommand{\gptm}{GPT-3$_{\text{350M}}$\xspace}
\newcommand{\gptb}{GPT-3$_{\text{1.3B}}$\xspace}
\newcommand{\gpthf}{GPT-2$_{\text{350M}}$\xspace}

%% file: _s0_abstract.tex
\begin{abstract}
Large-scale transformer models have become the de-facto architectures for various machine learning applications, e.g., CV and NLP. 
However, those large models also introduce prohibitive training costs. 
To mitigate this issue, we propose a novel random and layerwise token dropping method (\OURS), which skips the computation of a subset of the input tokens at all middle layers.
Particularly, \OURS achieves considerable speedups and comparable accuracy as the standard training baseline. 
Compared to other token dropping methods, \OURS does not require (1) any importance score-based metrics, (2) any   special token treatment (e.g., \texttt{[CLS]}), and (3) many layers in full sequence length training except the first and the last layers. 
Besides, a new \layertoken learning rate schedule is proposed for pretraining problems that resolve the heavy tuning requirement for our proposed training mechanism. 
Finally, we demonstrate that \OURS can be applied to broader applications, including \gpt and \bert pretraining as well as ViT and \gpt finetuning tasks. 
Our results show that \OURS can save about 33.3\% theoretical compute cost and 25.6\% wall-clock training time while achieving similar zero-shot evaluations on \gptb as compared to baseline.
\end{abstract}

%% file: _s1_intro.tex
\section{Introduction}
\label{sec:intro}

Large-scale transformers have been demonstrated to have supreme performance on natural language processing~\citep{tenney2019bert,radford2019gpt,colin2019t5}, computer vision~\citep{dosovitskiy2020image}, and other applications~\citep{gong2021ast,guo2021pct}. 
However, both the pretraining procedure and some downstream finetuning tasks (e.g., long document summary) are time-consuming and resource-hungry. 
Thus, there is a need to speed up the training and reduce the compute cost for large-scale transformer pretraining and finetuning.

Recently, \citet{hou-etal-2022-token}  adopt the token pruning/dropping/bypassing technique~\citep{kim2021learned,goyal2020power,kim2020length} from \bert inference to \bert pretraining by skipping the compute of part of the input tokens at some middle layers. 
The results of \citep{hou-etal-2022-token} (referred to as \tokenbypass) show that it can theoretically reduce the pretraining cost by 25\% for both \bertbase and \bertlarge without losing accuracy on finetuning tasks.
Although achieving great speedup, \tokenbypass  
(1) needs an import-score metric to determine the dropped tokens  and special token treatment to keep important tokens (e.g., \texttt{[CLS]}), both of which require manual designs;
(2) has to keep the first half layers and the last layer (in total, half of the depth) in full sequence length training, which limits its layer-bypassing ability.  
(3) solely focuses on \bert Masked-LM pretraining tasks and has not been applied to other tasks, e.g., causal-LM.
In this work,  we address those challenges and introduce our random and layerwise token-dropping method (\OURS). 
In summary, \emph{our contributions} are as follows:
\begin{itemize}[noitemsep, nolistsep, labelindent=0pt, leftmargin=*]
    \item All tokens are treated equally without any special token treatment or import-score measurement, i.e., no manual design, and are dropped in a purely random manner. 
    Meanwhile, instead of fully bypassing the dropped token for all middle layers~\citep{hou-etal-2022-token}, each layer in \OURS drops tokens independently from the other layers. 
    This helps the multi-head attention in the middle layers capture the dependency relation across different tokens suggested in~\citep{vig2019analyzing}.
    \item \OURS applies token dropping at all middle layers except the very first and last layers, which further reduces manual design and increases training efficiency. 
    We also propose a new monotonic sequence length growth method as training evolves to 
    (1) reduce the gradient noise introduced by \OURS for better convergence 
    and (2) close the training and inference (autoregressive generation) gap, since \OURS breaks the autoregressive manner in middle layers during training, for \gpt models. 
    \item To reduce the tuning effort for the newly proposed training procedure, we introduce a new \layertoken learning rate schedule, which scales the learning rate based on the sum of consumed tokens of each layer for pretraining tasks.\footnote{Note that the numbers of consumed tokens for different layers are different.}
    We show its superb performance for \OURS on \gpt/\bert pretraining compared to the standard iteration-based learning rate schedule. 
    \item We extensively test \OURS on both pretraining tasks, including \gpt and \bert pretraining, and finetuning tasks, including causal-LM finetuning for \gpt and image classification for \vit.
    For all tasks, \OURS achieves similar accuracy as the original baseline method with up to 33.3\% theoretical cost saving and up to 25.6\% wall-clock time saving. 
    \item Finally, we show that \OURS has a potential regularization effect, which can be used for both pretraining and finetuning problems. 
\end{itemize}

%% file: _s2_related_work.tex
\section{Background}
\label{sec:related_work}
Transformer~\citep{vaswani2017attention} architecture is a stack of transformer layers, each of which has two main ingredients, i.e., the multi-head attention (\mha) and the feed-forward connection network (\ffc). 
Suppose the transformer has $l$ layers denoted as $L_1,\ldots,L_{l}$.
Let $X_{i} \in \mathbb{R}^{s\times d}$ be the output tensor of $i-$th transformer layer, and $x_0$ the input (after embedding) of the transformer. 
Here $s$ is the sequence length and $d$ is the hidden dimension. 

Token dropping (or token bypassing/pruning)~\citep{kim2021learned,goyal2020power,kim2020length,press2021train,wang2021spatten} was originally proposed for \bert inference to reduce the computational overhead. 
In this case, if a token $i$ ($X_{j, i}$) is decided to be dropped at layer $j$ ($L_j$), the compute cost of this token through all remaining layers ($L_k$ where $k>j$) is eliminated. 
As such, the sequence length $s_i$ of the $i$-th layer's input $X_{i-1}$ will be a non-increasing array, i.e., $s_0 \geq s_1~...~\geq s_{l}$. 
However, such a configuration has been shown instability for adaptive token-dropping inference~\citep{kim2020length}.
Therefore, \citet{kim2020length} utilize the sandwich rule and  distillation from~\citep{yu2019universally} to stabilize training and boost accuracy.
But these two methods also significantly increase the training cost. 
Thus, such techniques cannot be applied to speed up the pretraining procedure.
Recently, \citet{hou-etal-2022-token} extended token dropping from inference to \bert pretraining (referred to as \tokenbypass). 
\citet{hou-etal-2022-token} use several importance scores/metrics to determine the dropped tokens, e.g., cumulative loss and frequency of each token. 
To overcome the training instability issue, the authors proposed two main mechanisms: 
(1) the sandwich token dropping rule, where the first (layer 1 to $i)$ and the last few layers (layer $L_{l-j}$ to $L_l$) of the \bert capture all tokens (i.e., no token dropping) and the middle layers bypass $s' \leq s$ tokens from $L_i$ to $L_{l-j}$.
Particularly, the authors (only) test on the encoder transformer (12-layer \bertbase and 24-layer \bertlarge), and let $i=l/2-1$, $j=1$, $s'=s/2$.
(2) special token treatment, where special tokens (e.g., \texttt{[MASK], [CLS], [SEP]}) are never dropped.

Compared to \tokenbypass from~\citep{hou-etal-2022-token}, our \OURS 
(1) does not require importance score metric, special token treatment, or the sandwich token dropping rule, which dramatically reduces the manual design effort;
(2) has been broadly tested on pretraining tasks, including \gpt and \bert, as well as finetuning tasks, including \vit classification and \gpt causal-LM. 
Meanwhile, we found out that directly applying \tokenbypass to causal-LM leads to severe accuracy degradation. 
Please see the detailed description of \OURS in~\sref{sec:methodology} and our extensive evaluation in~\sref{sec:results-pretraining} and~\ref{sec:discussion}. 
We also include a thorough discussion of other efficient training methods in~\appref{sec:other_efficient_training_approaches}.

%% file: _s3_method.tex
\section{Methodology}
\label{sec:methodology}
\subsection{Random and Layerwise Token Dropping Method}
\label{subsec:rand-ltd-main}
\textbf{Layerwise Token Dropping Mechanism.}
As pointed out in~\sref{sec:related_work}, existing inference and training token dropping methods either permanently drop tokens from the compute graph at intermediate layers, or at least make some tokens fully skip a consecutive series of middle layers. 
However, several works~\citep{vig2019analyzing,michel2019sixteen,voita2019analyzing} have shown that \mha focuses on different tokens at different layer depths and the attention map aligns with the dependency relation most strongly in the middle of transformer architectures. 
Therefore, \tokenbypass used in~\citet{hou-etal-2022-token}, i.e.,  fully skipping middle layers, may hinder the learnability/generalization of the architecture during pretraining/inference. 
We conjecture that this might be why multiple first/last layers need to be kept and the special token treatment is needed in~\citep{hou-etal-2022-token}. 
To further verify if this fully skipping middle layer mechanism~\citep{hou-etal-2022-token} causes any learnability issue, we apply \tokenbypass on \gpt finetuning tasks and observe much lower performance as compared to baseline.
See more details in~\sref{sec:layerwise_vs_tokenbypass}.

In order to overcome this problem, we now propose a layerwise token dropping (\ltd) mechanism. 
Instead of fully bypassing dropped tokens over all middle layers, each transformer layer independently drops/retains its own set of tokens.
In more detail, recall that the input of $(i+1)$-th layer ($L_{i+1}$) is $X_{i}\in\mathbb{R}^{s\times d}$. 
Denote the dropped token index as $J_i=\{j_1, j_2, ..., j_{a_i}\}$ and the kept token index as $K_i=\{k_1, ..., k_{b_i}\}$ such that $a_i+b_i=s$.
We have $J_i\cup K_i = \{1,2,3...,s\}$ and $J_i\cap K_i=\emptyset$ for each layer.
Meanwhile, for any two different layers $L_{i_1}$ and $L_{i_2}$, $J_{i_1}$ and $J_{i_2}$ are independent, though the dropped ratios are the same.
With this layerwise mechanism, each token rarely bypasses all middle layers. Thus, its dependency on other tokens can be captured by \mha. 

\textbf{Random Token Dropping.}
Various important score-based metrics are used to determine the token dropping criterion. 
Most of them can be categorized in two ways: attention score related metrics or loss/frequency-based metrics. 
However, both of them introduce challenges that make LTD less practical. 
Particularly, for attention score-based metrics, the compute cost for \ltd is too high since the metric has to be calculated for every layer; 
for loss-/frequency-based metrics, generally accumulated loss or frequency is used and this accumulated metric would not be changed within the same iteration (a.k.a. one forward pass of the network). 
Therefore, the unchanged loss/frequency metric leads the dropped token to be the same for different layers, making the token dependency not be captured by the \mha of middle layers~\citep{vig2019analyzing}.

To satisfy the independent requirement of \ltd, we propose to use \emph{purely random} token dropping assignment. 
For each transformer layer, we randomly (uniformly) select a small batch of tokens to proceed with the compute and drop the rest. 
In more details, assume $M_i=$\{$m_i(1)$, $m_i(2)$, ..., $m_i(s)$\} is a random shuffle of $S=$\{1, 2, ..., s\}. 
Then the dropped token set is $J_i=$\{$m_i(1)$, $m_i(2)$, ..., $m_i(a_i)$\} for the input of $L_{i+1}$.

\textbf{Random and Layerwise Token Dropping.}
Combining layerwise token dropping with random token dropping, we have our final random and layerwise token dropping method (\OURS), 
which can efficiently apply token dropping for each individual layer and 
can capture the attention dependency of each token with other others in middle layers with high probability.

\begin{figure}
\begin{minipage}{.55\linewidth}
  \includegraphics[width=0.9\linewidth]{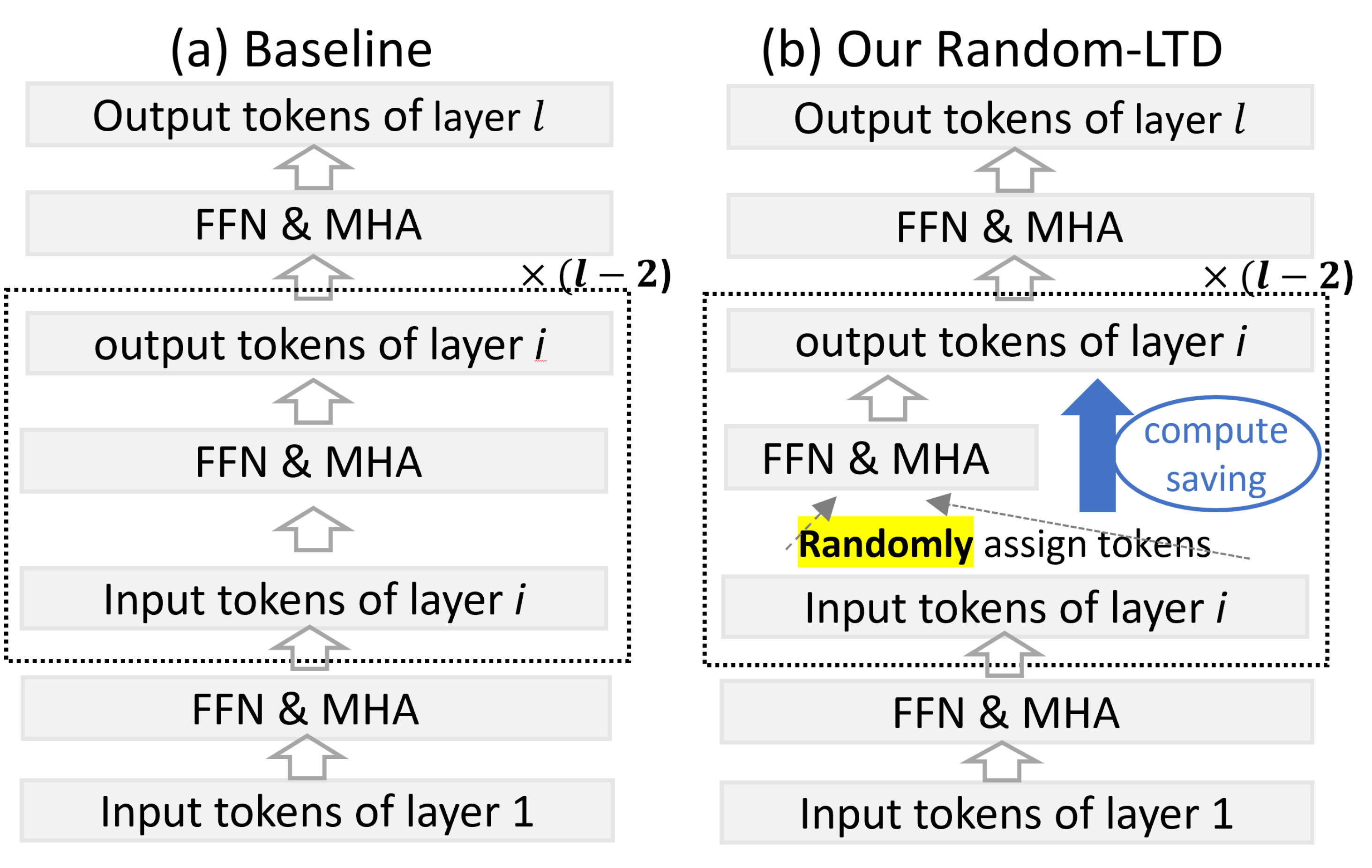}
  \vspace{-0.3cm}
    \caption{Transformer layers for baseline  and \\\OURS training. The dash-line  box is \\ repeated by $l-2$ times (see more in \fref{fig:illustration_of_random_ltd_and_baseline_and_tokenbypass}).}
    \label{fig:illustration_of_random_ltd_and_baseline}
\end{minipage}
\begin{minipage}{.445\linewidth}
\begin{lstlisting}
if meth == "baseline":
  hs = Layer(hs) 
if meth == "|\color{upforestgreen}{random-LTD}|":
  k_hs, d_hs = |\color{upforestgreen}{gather}|(hs)
  k_hs = Layer(k_hs)
  hs = |\color{upforestgreen}{combine}|(k_hs, d_hs)
\end{lstlisting}
 \caption{\OURS only requires a few lines of code. 
 Here hs, k$_{\text{hs}}$, and d$_{\text{hs}}$ means the full input, kept input, and dropped input. 
 ``gather'', ``Layer'', ``combine'' means the functions for random selection, transformer layer, and token combination.
 }
\label{fig:code_of_random_ltd_baseline}
\end{minipage}
\vspace{-0.5cm}
\end{figure}

The illustration of the comparison between standard baseline training and \OURS is shown in ~\fref{fig:illustration_of_random_ltd_and_baseline} (an additional comparison with~\citep{hou-etal-2022-token} in  \fref{fig:illustration_of_random_ltd_and_baseline_and_tokenbypass}).  The pseudo-code is given in~\fref{fig:code_of_random_ltd_baseline}.
For each layer, as compared to the baseline, \OURS randomly selects (function ``gather'' in~\fref{fig:code_of_random_ltd_baseline}) a subset of the tokens and feeds (function ``Layer'' in~\fref{fig:code_of_random_ltd_baseline}) them into the transformer layer. 
Afterward, we combine (function ``combine'' in~\fref{fig:code_of_random_ltd_baseline}) the output of transformer layer with the dropped tokens to recover the full sequence length.
Thus, the next layer still receives the full sequence and can repeat this process.

Since the dropped tokens of each layer are independent, there is no need for \OURS to treat special tokens (e.g., \texttt{[MASK], [CLS], [SEP], [PADDING]}) differently from other normal tokens, which can further reduce the cost of computing the dropping criterion. 
Meanwhile, we show that special token treatment does not bring extra benefits for \OURS on \bert pretraining in~\sref{sec:no_need_special_token_treatment}.

\subsection{Dropping Schedule of \OURS}
\label{subsec:rand-ltd-schedule}
\textbf{Layers without Token Dropping.}
While \tokenbypass~\citep{hou-etal-2022-token} needs to keep half of the layers in full sequence length training, \OURS has no such limitation.
Thanks to the attention-capture feature of \OURS, we can apply \OURS to most of the transformer layers except the first and last transformer layers. 

Keeping the first and last layers in full sequence length training usually leads to better performance since 
(1) the first layer directly connects to the embedding, and it can help refine the raw feature;
(2) the last layer directly connects to the final prediction; a feature realignment for all tokens can improve the model quality. 
We also provide a detailed study to show the importance of keeping the first and last layers without token dropping in~\sref{sec:why_we_need_keep_first_last_layer}.

\textbf{Monotonic Sequence Length Growth.}
 
In order to reduce the gradient variance introduced by \OURS for better training, we monotonically increase the kept sequence length throughout training (referred to as \pslg) with a linear schedule. 
Particularly, the dropped token set $J_i$ for the $i$-th layer gradually shrinks and the kept token set $K_i$ gradually grows as the training proceeds. 
Denote the size of $J_i$ ($K_i$) at step $t$ is $a_{i, t}$ ($b_{i, t}$), its final size is $0$ ($s$), and the total training iterations is $T$. 
Assume we want to gradually reduce the size of $J_i$ to zero at iteration $T'$ and the decreasing strength is $s_{dec}$. 
Then the decreasing step size is 
    $T_{dec} = T' /({a_{0, t}}/{s_{dec}})$,
i.e., for every $T_{dec}$ iterations, the size of $J_{i}$ ($K_i$) reduces (increases) by $s_{dec}$.
Please see~\fref{fig:gpt_pretraining_main_fig} for an illustration of $K_{i}$ on \gpt pretraining.
We also show that \pslg outperforms the constant drop schedule with similar compute savings in~\sref{sec:why_we_need_progressve_seq_len_growth}.

\subsection{New Learning Rate Schedule for Pretraining}
\label{sec:learning_rate_schedule_for_pretraining}
When performing pretraining on language models, we oftentimes use a decaying learning rate schedule based on iteration with a warmup period. 
Particularly, at the first few thousand or hundred iterations, warming up the learning rate is critical for distributed pretraining tasks due to its instability~\citep{goyal2017accurate,li2021curriculum}. 
However, an iteration-based schedule is not optimal for \OURS.  

First, \OURS reduces the effective batch size of middle layers at the initial warmup phase. 
The effective training tokens for dropped token layers become much smaller than the baseline training.
Second, for most of our training cases, \pslg does not reach the full length until $>2/3$ of training iterations for large compute saving. 
At such time, the iteration-based learning rate is considerable small. 
And this small learning rate cannot provide efficient training dynamics for \OURS. 
Therefore, to stabilize the initial training phase and to have a large enough learning rate in the later training phase, we need to increase the warmup iterations and slow down the learning rate decay. 
Here, we propose a new learning rate schedule based on the layerwise tokens consumption, called layer-token learning rate (\layertokenlr). 
Please see~\appref{sec:formal_layertokenlr_description} for the formal and detailed description of \layertokenlr. 

We emphasize that one can always tune the learning rate schedule by increasing the maximum learning rate or the warmup iterations. 
However, it would require a lot of engineering effort. 
Therefore, we propose this \layertokenlr schedule, which is more suitable for our \OURS than the standard one. 
We also include a detailed comparison between the standard learning rate schedule and \layertokenlr in~\sref{sec:lr_schedule_effect}.

%% file: _s4_results.tex
\section{Main Results}
\label{sec:results-pretraining}

In this section, we first provide the results of \OURS for pretraining on \gpt and \bert models. 
We then extend \OURS on the computer vision domain to demonstrate its broader applications.    
Similar to~\sref{sec:learning_rate_schedule_for_pretraining} and~\appref{sec:formal_layertokenlr_description}, we use the \layertoken compute the cost to measure the total training budget.\footnote{Similar to~\citep{hou-etal-2022-token}, we do not include (1) the final prediction layer and (2) the attention compute difference between the different lengths of sequence for the compute cost comparison.
}
We also provide the real training time saving for \gpt and \bert pretraining.
Kindly note that the real-time saving depends on various factors, e.g., the implementation and hardware. 

\begin{wrapfigure}{r}{0.6\linewidth}
  \vspace{-0.5cm}
    \centering
    \includegraphics[width=1.0\linewidth]{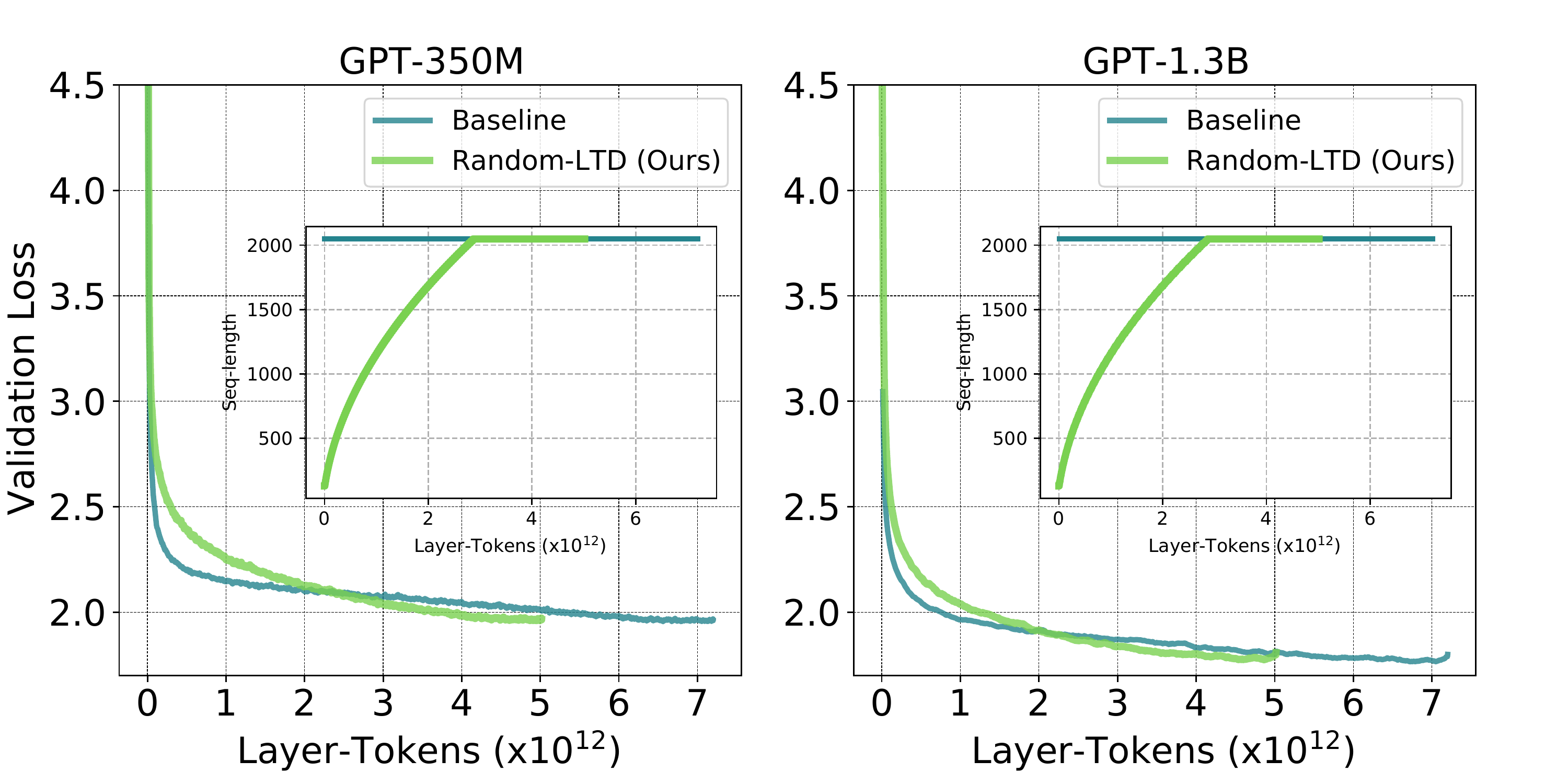}
      \vspace{-0.5cm}
    \caption{The comparison of validation curves between baseline and \OURS on \gpt pretraining. 
    Here the x-axis is based on \layertoken consumption. 
    The sequence length is illustrated in the inserted figure.}
    \label{fig:gpt_pretraining_main_fig}
    \vspace{-0.5cm}
\end{wrapfigure}

\subsection{\gpt pretraining}
We train \gpt-3-style models with 350 million parameters (\gptm) and 1.3 billion parameters (\gptb) on PILE dataset~\citep{gao2020pile} and the total number of training tokens is 300 billion. 
For \OURS, the initial dropped token length for all middle layers is 1920 (i.e., 128 tokens are kept for compute), and it decreases by 16 for every 1.75B training tokens. 
After 210B training tokens, \OURS degrades to standard training procedure with full sequence length. 
Theoretically, this can save 1/3 of the \layertoken training budget. 
See~\appref{sec:gpt_experimental_setup} for more details.

The evaluation loss curves for baseline and \OURS are shown in~\fref{fig:gpt_pretraining_main_fig}.
As can be seen, for both \gptm and \gptb, \OURS has similar evaluation losses as the baseline with 1/3 less \layertoken consumption.
We also provide the zero-shot evaluation results in~\tref{tab:gpt3_main_result}. 
For both \gptm and \gptb, \OURS achieves comparable results as the baseline, 
Besides, \OURS can save 14.3\% wall-clock training time on \gptm and 25.6\% wall-clock training time on \gptb. 

We reiterate that the \layertoken consumption saving ratio cannot directly transfer to GPU wall-clock training time saving ratio due to the implementation/hardware. 
Meanwhile, note that the saving number we reported here is not the maximum potential saving (with fixed implementation/hardware, etc) since we can reduce the training GPU numbers for \OURS at the initial training phase, which has a shorter effective training sequence length.
Also, although \OURS has the same theoretical compute saving for both \gptm and \gptb, the real wall-clock time saving varies a lot
because \gptb has a larger hidden dim size, which means the model spends more time on real computing than other operators, e.g., data movement and gradient communication.

\begin{table}[!htb]
\begin{minipage}[t]{.48\linewidth}
    \caption{
Zero-shot evaluation results (last col.) on \gptm and \gptb. 
The GPU cost provided here is the used number of A100-40G times the training days.
See~\tref{tab:gpt3_main_result_full} for 19 tasks.
}\centering
\label{tab:gpt3_main_result}
\begin{adjustbox}{width=1\linewidth}
\centering
\setlength\tabcolsep{1pt}
\begin{tabular}{lcccccccccccccc }
\toprule
Model & Method &\layertoken Saving & GPU Cost (saving)   & Ave. \\
\midrule
\multirow{2}{*}{\gptm} & baseline & None &64$\times$2.59 (0.0\%) & 38.6 \\
                       & \OURS   & 33.3\% &64$\times$2.22 (14.3\%) & 38.9 \\
\cdashlinelr{1-15}
\multirow{2}{*}{\gptb} & Baseline& None & 64$\times$5.42 (0.0\%) & 42.7 \\
                       & \OURS   & 33.3\% & 64$\times$4.03 (25.6\%) & 42.5 \\
\bottomrule
\end{tabular}
\end{adjustbox}
\end{minipage}\hfill
\begin{minipage}[t]{.5\linewidth}
\caption {Finetuning results for \bertlarge.  
The training cost provided here is the used number of A100-40G times the training days. 
See~\tref{table:bert-std} for full results.
}\label{table:bert-main}
\begin{adjustbox}{width=0.99\linewidth}
\centering
\setlength\tabcolsep{2pt}
\begin{tabular}{lccccccccccc}
\toprule
  Method    & \layertoken Saving      &   GPU Cost (Saving) & Ave.  \\\midrule
baseline  & None  & 64$\times$5.89 (0.0\%)    & 85.42 \\
\cdashlinelr{1-9}
\OURS-1  & 26.2\% & 64$\times$5.42 (7.95\%)   & 86.95 \\
 \OURS-2  & 31.1\% & 64$\times$5.21 (11.5\%)  & 86.42\\
\bottomrule
\end{tabular}
\end{adjustbox}
\end{minipage}\hfill
\end{table}

\subsection{\bert Pretraining}\label{subsec:bert-pretraining}
 
We pretrain \bertlarge on PILE dataset for 2M iterations with batch size 1024 and sequence length 512 following~\citep{shoeybi2019megatron}.
We apply \OURS with two variants, \OURS-1 and \OURS-2. 
Particularly, for \OURS-1 (\OURS-2), the initial kept token length is 200 (128), and it increases by 16 for every 48B (38B) training tokens. 
As such, we save 26.2\% (31.1\%) \layertoken consumption for \OURS-1 (\OURS-2).
We evaluate the trained model on four downstream tasks as~\citep{shoeybi2019megatron}, i.e., MNLI, QQP, RACE-m, RACE-h. 
Note that we apply standard finetuning without token dropping to have a fair comparison for both pretrained models from the baseline and \OURS. 
Please see~\appref{sec:bert_experimental_setup} for more details. 

\tref{table:bert-main} summarizes the results along with the full results in~\tref{table:bert-std}. 
Although \OURS is slightly worse than baseline on a certain task (QQP, see~\tref{table:bert-std}), it gives much higher accuracy on other tasks while saving 26.2--31.1\% of the theoretical computation overhead in pretraining. 
Overall, \OURS-1 achieves 1.54 points higher average accuracy over baseline and \OURS-2 achieves 1 point higher average accuracy over baseline.

Meanwhile, \OURS-1 (\OURS-2) saves about 7.95\% (11.5\%) wall-clock time as compared to baseline. 
Note that similar to \gpt pretraining, the saving depends on the implementation/hardware, and \OURS has potentially larger savings if elastic training is performed. 
Also, although \gptm and \bertlarge have similar model sizes as well as similar theoretical compute saving, the final wall-clock training time saving varies by about 3\%. 
This is caused by \bertlarge having a shorter final sequence length  (i.e, 512) than \gpt (i.e, 2048), and the real compute time for a sentence with sequence length 128 is not 1/4 (or 1/16) of a sentence with 512 (2048) tokens. 
This leads the overall compute time-saving for \gptm to be larger than that for \bertlarge.

\subsection{\vit Finetuning}
\label{sec:vit-fine-tuning}

We perform the vision transformer (\vit) on both ImageNet (with a 12-layer pretrained \vit) and CIFAR (with a 24-layer pretrained \vit).
For \OURS, the initial sequence length is 66 and linearly reaches the 197 full sequence length at 80\% of the total training iterations such that  22.3\% layer-token saving is achieved. See training details in \appref{subsec:fine-tune-vit}. 
We summarize the result with standard deviation in~\tref{table:vit-main} along with the full details in~\tref{table:vit-main-full}. 
As can be seen, \OURS can achieve comparable results as the baseline on all three datasets. 
This demonstrates the broader applications of \OURS.

\begin{table}[!htb]
\vspace{-0.4cm}
\begin{minipage}[t]{.48\linewidth}
\caption{Finetuning result of \vit on ImageNet.
See~\tref{table:vit-main-full} for the results on CIFAR10/100.
} \label{table:vit-main}
\begin{adjustbox}{width=0.99\linewidth}
\begin{tabular}{lcccccc}
\toprule
           & \multicolumn{3}{c}{ImageNet datasets on 12-layer ViT}            \\
Method &\layertoken Saving & Top-1       & Top-5         \\\midrule
baseline   & N/A            & 84.65±0.04 & 97.41±0.02 \\
random-LTD & 22.3\%      & 84.70±0.04  & 97.48±0.02  \\\bottomrule
\end{tabular}
\end{adjustbox}
\end{minipage}\hfill
\begin{minipage}[t]{.5\linewidth}
\caption{Ablation study of special token treatment for \bert pretraining with 22.2\% \layertoken saving. 
See~\tref{table:bert-ablation-study-token-full} for all results.
}
\label{table:bert-ablation-study-token}
\begin{adjustbox}{width=0.99\linewidth}
\centering
\begin{tabular}{ccccccccc}
\toprule
  Keep Special Tokens  & Pretraining PPL val/test  & Downstream Ave.  \\\midrule

yes      & 6.024 / 6.049& 88.50\\
 no      & 6.018 / 6.040 & 88.52 \\
 \bottomrule
\end{tabular}
\end{adjustbox}
\end{minipage}\hfill
\end{table}

%% file: _s5_discussion.tex
\section{Discussion}
\label{sec:discussion}
In this section, we present several import ablation studies and the potential regularization effect of \OURS. 
Besides the three tasks used in previous sections, we also include \gpt finetuning on causal-LM problems using the \gpthf from Huggingface~\citep{wolf2019huggingface}. 
Please see~\appref{subsec:fine-tune-gpt} for the training details.
Also, we reduce the iterations of \bert pretraining from 2M to 200k due to resource limitations. 
Please see~\appref{sec:bert_experimental_setup} for more details.

\subsection{Layerwise Token Dropping vs. \tokenbypass}
\label{sec:layerwise_vs_tokenbypass}
Although \tokenbypass~\citep{hou-etal-2022-token} demonstrates its great ability on \bert pretraining, its skipping policy may still hurt the performance of other tasks, e.g., causal-LM. 
The reason is mentioned in~\sref{sec:methodology}, i.e., \mha of middle layers focuses on different tokens at different depths. 
Fully skipping those layers may lead the causal-LM task to lose attention capability. 
However, \OURS does not have this issue as it randomly selects kept tokens for each layer. 

To verify this, we provide an ablation study on the comparison between  \OURS and \tokenbypass with \gpthf finetuning on PTB~\citep{marcus-etal-1993-building}.  
We make two sets of experiments:
\begin{itemize}[noitemsep, nolistsep, labelindent=0pt, leftmargin=*]
    \item
    \textbf{Set 1.} Following~\citep{hou-etal-2022-token}, we bypass half of the tokens based on their empirically moving average loss from $L_{12}$ to $L_{23}$. 
    Similarly, we apply \textit{constant} random token drop to the layers from the middle to the second last ($L_{12}$ to $L_{23}$). 
    \item
    \textbf{Set 2.} 
    We apply \tokenbypass or constant random token dropping for half of the tokens starting from the second layer ($L_2$) until the second last layer ($L_{23}$). 
\end{itemize}
The validation curves of two cases are shown in~\fref{fig:layerwise_vs_tokenbypass}. 
As can be seen, for both cases, \OURS performs much better  than \tokenbypass. 
Particularly, for the Set 2 comparison, the perplexity of \OURS is about 10 points lower than \tokenbypass, demonstrating that \OURS can be applied to more layers than \tokenbypass.

\begin{minipage}{0.48\linewidth}
    \begin{figure}[H]
    \vspace{-0.5cm}
    \centering
      \includegraphics[width=1.0\linewidth]{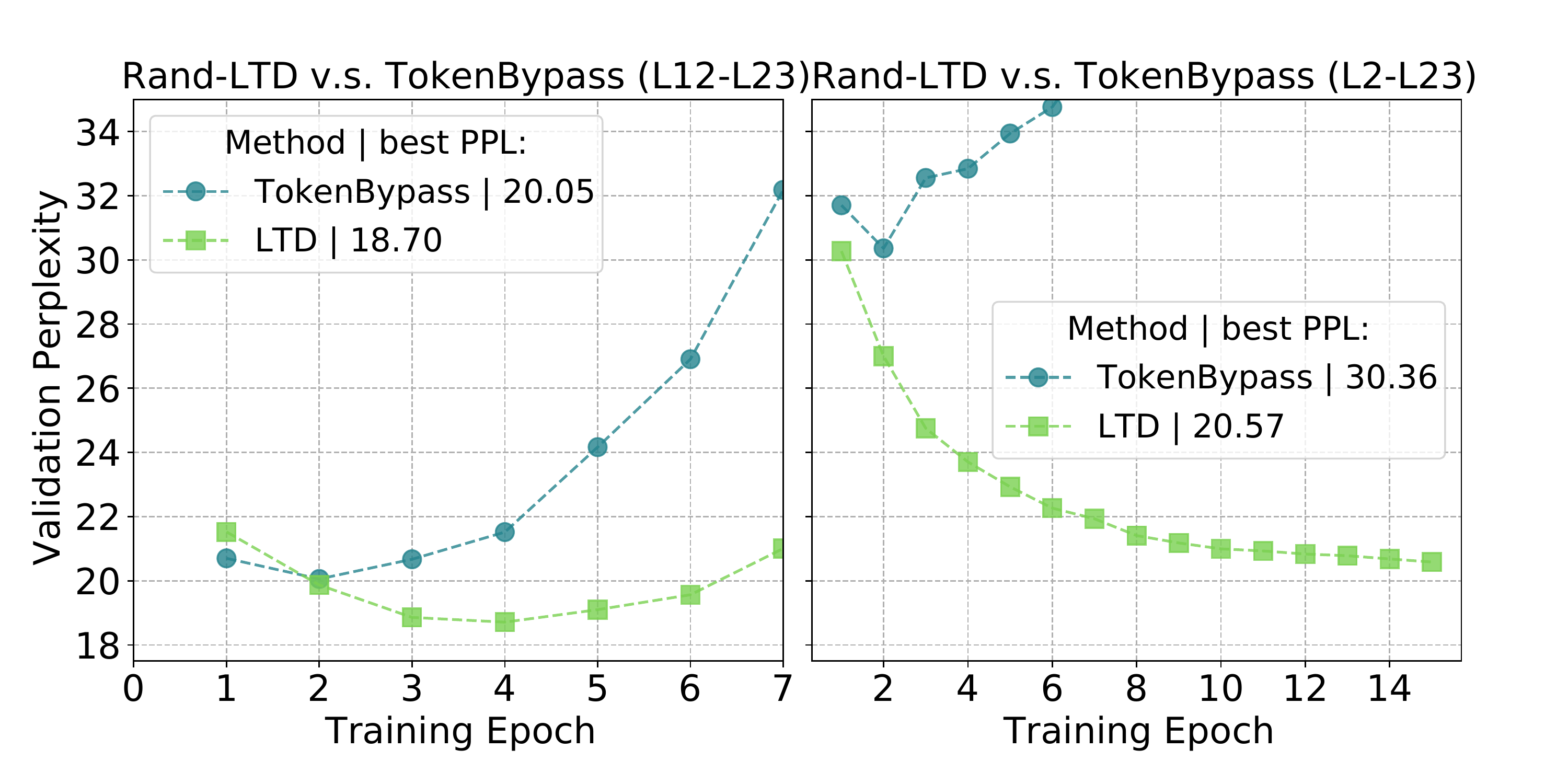}
    \vspace{-0.54cm}
    \caption{Comparison between random  \\ token dropping and \tokenbypass. 
 }
    \label{fig:layerwise_vs_tokenbypass} 
\end{figure}
 \end{minipage}
  \begin{minipage}{.485\linewidth}
\begin{figure}[H]
\vspace{-0.5cm}
    \centering
      \includegraphics[width=1.\linewidth]{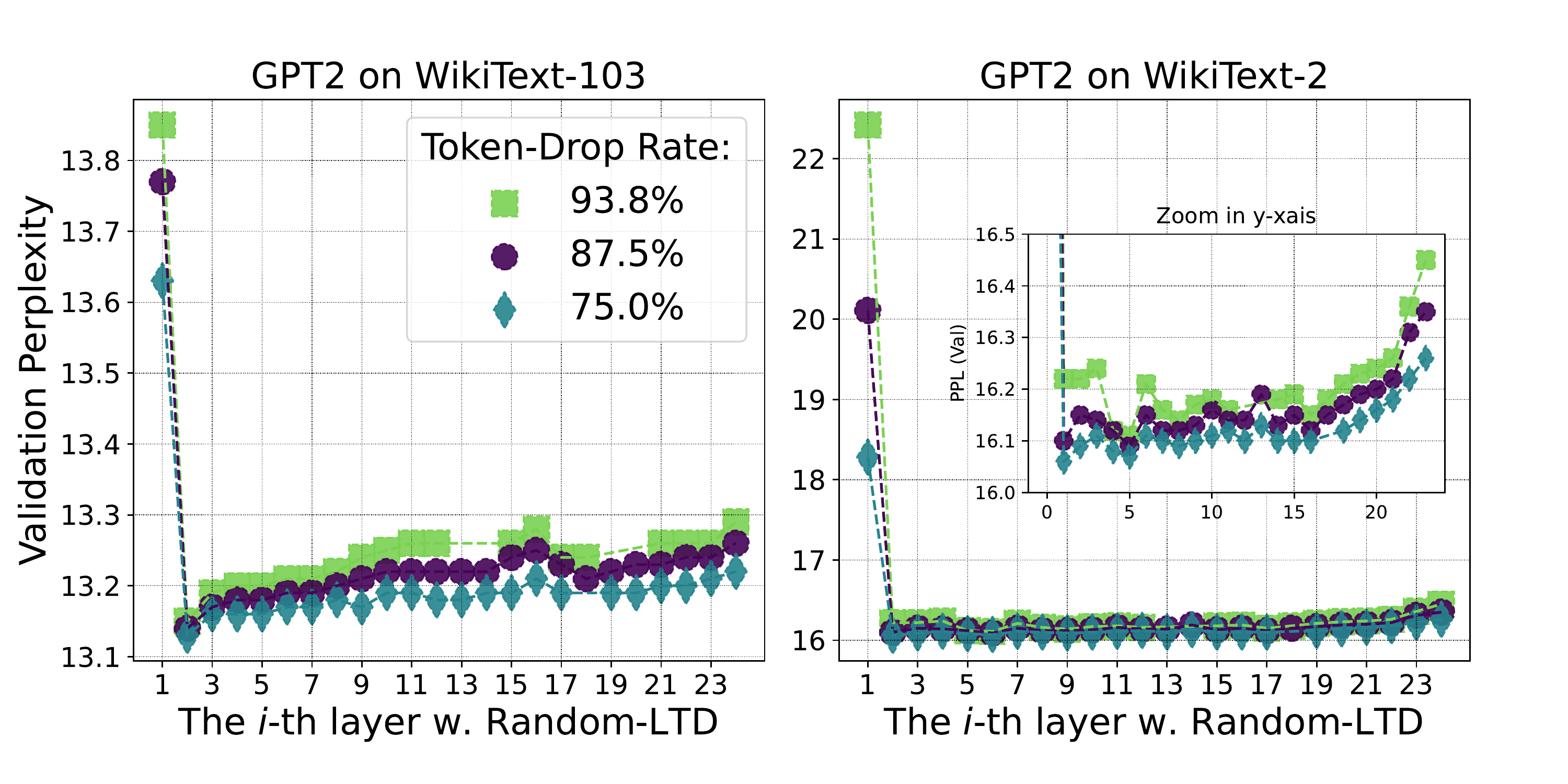}
   \vspace{-0.54cm} \caption{Sensitivity of random token dropping of each layer.
   }
    \label{fig:sensitivity}
\end{figure}
  \end{minipage}

\subsection{With/Without Special Token Treatment}
\label{sec:no_need_special_token_treatment}

Different from the \gpt pretraining task, which has consecutive sentences/paragraphs, the BERT pretraining data consists of two sentences that could be unrelated. 
The special tokens \texttt{[CLS]} and \texttt{[SEP]} play critical roles for the model in determining the beginning/end of each sentence so that the model can predict the relationship between the two sentences. 
Thus, there could be potential gain in keeping the tokens for all layers, which has a more detailed discussion in~\citep{hou-etal-2022-token}. 

Here we present an ablation study on whether keeping those special tokens helps \OURS or not. 
We perform a straight comparison for \OURS: (1) one with purely random selection and (2) the other with an additional criterion, i.e., keeping the special tokens for all layers. 
See training details in~\appref{sec:bert_experimental_setup}

The results of MNLI/QQP are shown in~\tref{table:bert-ablation-study-token} along with the full details in~\tref{table:bert-ablation-study-full}.
As can be seen, for both pretraining loss and downstream finetuning, special token treatment does not provide any benefit for \OURS.
Note that \OURS without special token treatment is also more compute-friendly for token dropping. 

\subsection{Why we need to keep the first and last layer?}
\label{sec:why_we_need_keep_first_last_layer}

To understand why keeping the first and last layers in full sequence length training, we present single layer sensitivity analysis shown in~\fref{fig:sensitivity} for \gpthf finetuning on Wikitext-2 and Wikitext-103. 
Particularly, we apply constant token dropping for one layer and keep all the other layers in the standard training mode. 
After the training, we measure the PPL and use it as the sensitivity metric, i.e., higher PPL indicates high sensitivity and vice versa. 
The U-shape of both curves  implies that the first and last ones are most sensitive to token dropping. 

\begin{table}[!htb]
\vspace{-0.5cm}
\begin{minipage}[t]{.55\linewidth}
\caption{Comparison of applying \OURS to \\ different layers on \gpthf finetuning and \vit\\ finetuning.
See~\tref{table:main-layers-full} with standard deviation.} 
 \label{table:main-layers}
\centering
 \begin{adjustbox}{width=0.99\linewidth}
\begin{tabular}{lcccccccc}
\toprule
   &  &\multicolumn{4}{c}{Apply \OURS except for Layer}        \\
  Metric  & dataset & None & First &  Last & First\&Last                     \\
\midrule 
\multirow{3}{*}{Perplexity} & PTB  & 16.00 & 16.01  & 16.09 & 15.92    \\
 &WikiText-2  &  17.06 & 17.01 & 17.01 & 16.94    \\
 &WikiText-103  &13.27  &13.03 & 13.23 &12.99     \\
 \midrule
\multirow{1}{*}{Accuracy}  & ImageNet-Top1  & 84.47 & 84.51 & 84.65  & 84.70     \\
 \bottomrule
\end{tabular}
\end{adjustbox}
\end{minipage}\hfill
\begin{minipage}[t]{.445\linewidth}
\caption{
Compare between \pslg and constant token dropping schedules. 
See~\tref{table:seq-schedules-full} for the full result with standard deviation and the result on \vit finetuning.
}
\setlength\tabcolsep{1pt}
\label{table:seq-schedules}
 \begin{adjustbox}{width=0.99\linewidth}
\begin{tabular}{l|cccc|ccc}
\toprule
dataset      & \multicolumn{3}{c}{PTB  (Metric: perplexity)} \\
Token-drop schedules   &   constant & constant  & \pslg\\
\layertoken saving    & 23.0\%   & 32.1\%      & 33.7\%       \\\midrule
Performance & 18.27   &     20.76        &   15.92  \\
\bottomrule
\end{tabular}
\end{adjustbox}
\end{minipage}\hfill
\vspace{-0.3cm}
\end{table}

To  further understand if \OURS can be applied to all layers when using \pslg, we include other three scenarios, i.e., applying \OURS to (1) all but not last layer, (2) all but first layer, and (3) all layers. 
We perform finetuning tasks on both causal-LM and image classification. 
See the full training details in~\appref{subsec:fine-tune-vit} and~\appref{subsec:fine-tune-gpt}.  
From~\tref{table:main-layers} and~\ref{table:main-layers-full}, we can clearly see that keeping the first and the last layers intact leads to a substantial improvement (beyond standard deviation) over the rest three scenarios.

\subsection{Why we need sequence length growth?}
\label{sec:why_we_need_progressve_seq_len_growth}

We now give an ablation study on why \pslg schedule is necessary. 
Again, We perform finetuning tasks on both causal-LM and image classification. 

We specially set a constant token dropping rate that matches the token saving of \pslg schedules with all other hyperparameters fixed.
We present the results in~\tref{table:seq-schedules} and~\ref{table:seq-schedules-full}. 
It can be clearly seen that given the almost same amount of \layertoken saving ($33\%-35\%$), the constant dropping schedule has worse performance than \pslg. 
\pslg schedule can actually be even better or comparable to those constant ones whose saving is $10\%$ smaller.

\subsection{\layertokenlr Schedule effect}
\label{sec:lr_schedule_effect}
To study the effectiveness of \layertokenlr, we compare three training scenarios for \gptm with 300B training tokens (see \appref{sec:gpt_experimental_setup} for training details): (1) the baseline training with the standard learning rate, (2) \OURS with the standard learning rate, and (3) \OURS with \layertokenlr. 
The validation curves and their corresponding learning rates with respect to iterations are plotted in~\fref{fig:token-lr-gpt}. 
As can be seen, the green curve (\OURS with \layertokenlr) can achieve comparable validation loss as the baseline, which is better than \OURS with the standard learning rate. 
This confirms that the small learning rate introduced by the standard learning rate schedule slows the learning of \OURS at the later training phase. 
A similar observation is made for the BERT pretraining, which will be deferred to~\appref{sec:bert-lr} due to the space limit.

\begin{minipage}{0.485\linewidth}
    \begin{figure}[H]
    \centering
    \includegraphics[width=1.0\linewidth]{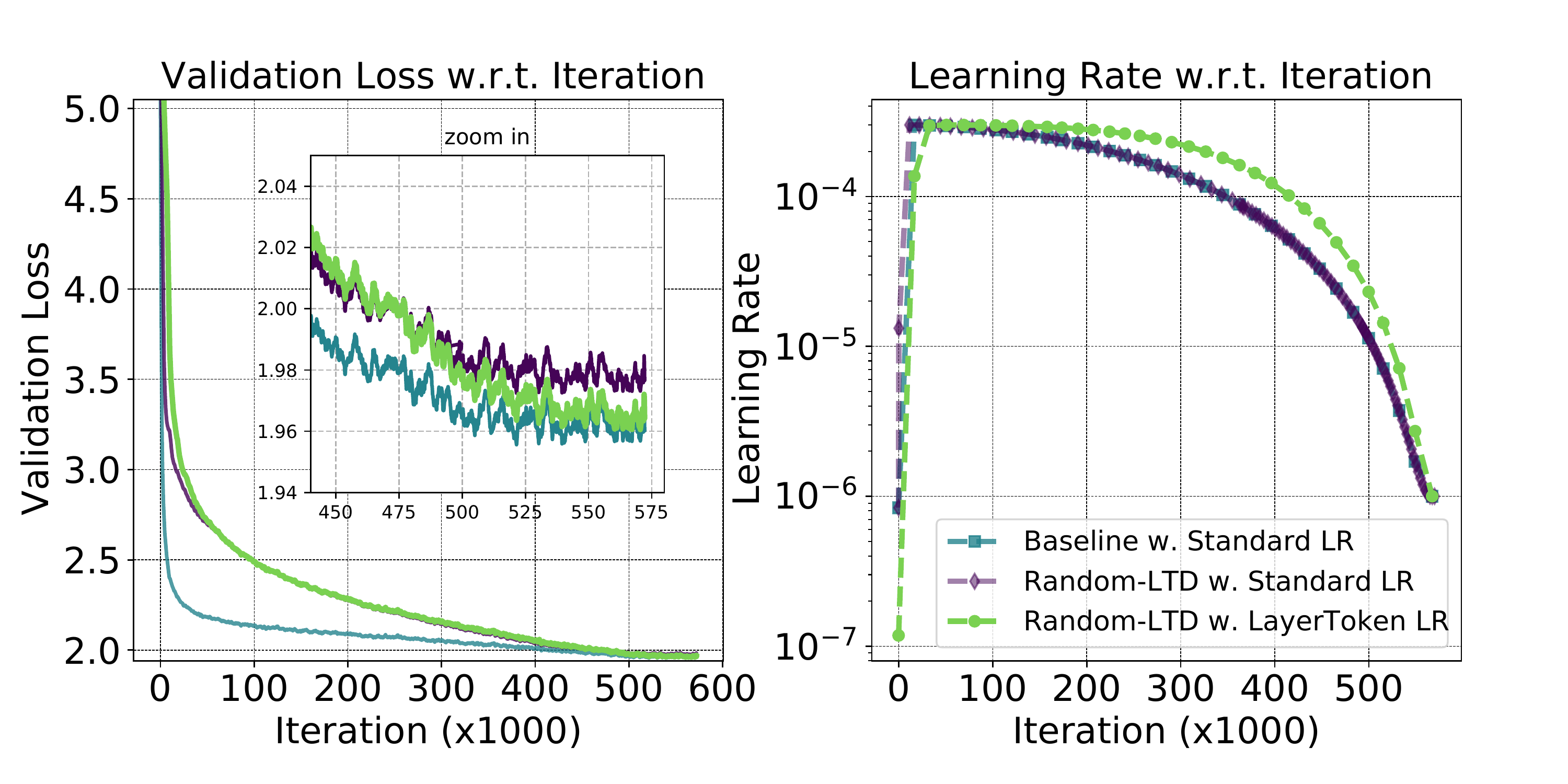}
     \vspace{-0.5cm}
    \caption{Comparison between a standard \\learning rate and our proposed learning rate\\ based on layerwise tokens}
    \label{fig:token-lr-gpt}
   
\end{figure}
 \end{minipage}
  \begin{minipage}{.485\linewidth}
\begin{figure}[H]
    \centering
      \includegraphics[width=1.\linewidth]{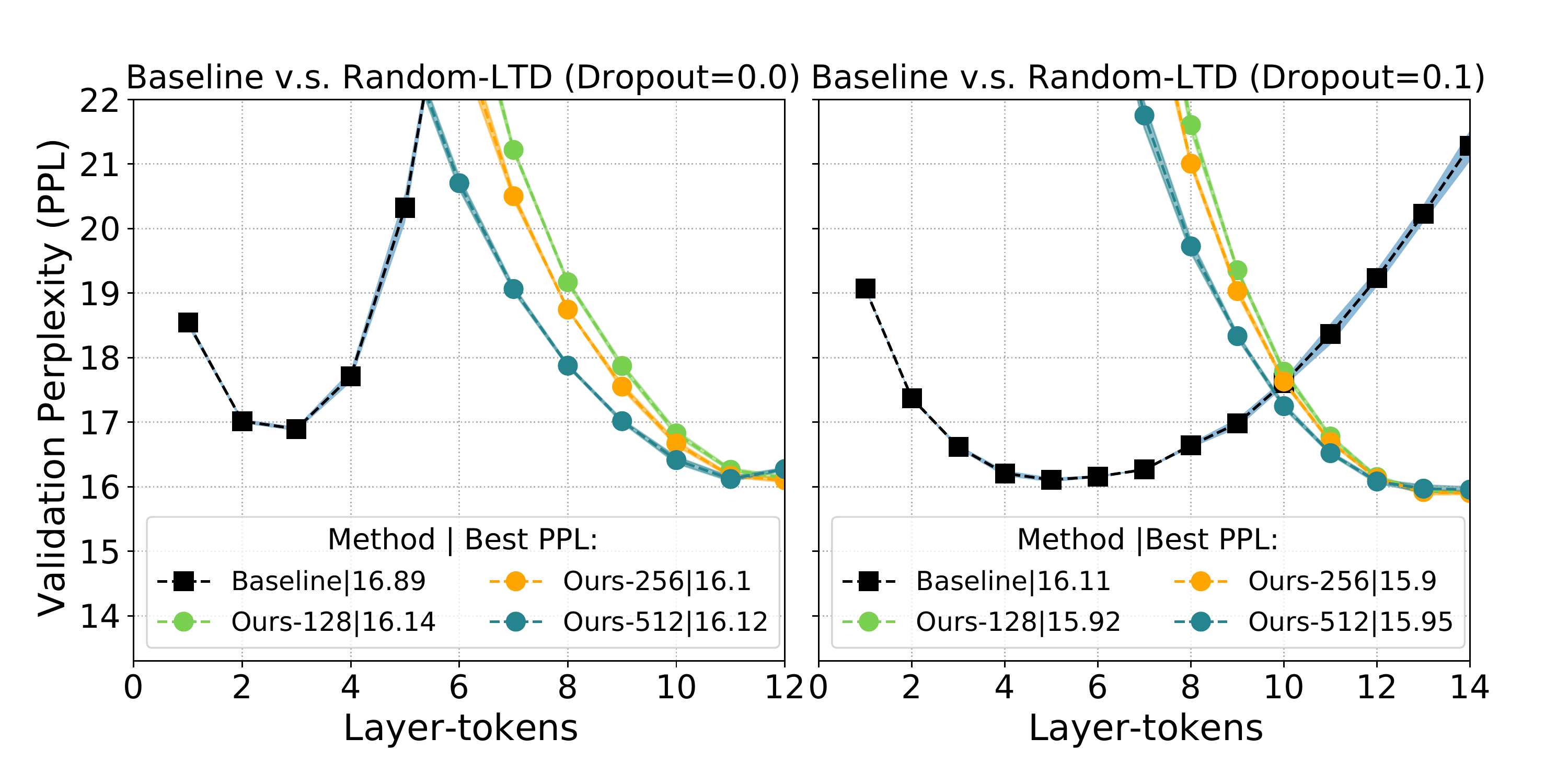}
     \vspace{-0.5cm}
    \caption{Study the regularization of \OURS on \gpthf finetuning on the PTB task without dropout (left) and with dropout (right).}
    \label{fig:regularization} 
\end{figure}
  \end{minipage}

\subsection{Comparison on \gptb with Various Training Budgets}
In this section, we perform various training budgets to train \gptb and compare the performance of baseline and \OURS to verify if \OURS can consistently save the training cost.

To fully understand the effectiveness of \OURS,  we train \gptb using a baseline with 120B (i.e., 2880 billion \layertoken consumption) to 360B tokens (i.e., 8640 billion \layertoken consumption).
For \OURS, we follow our~\sref{sec:results-pretraining} setting to save 1/3 of the total training budget and apply it to three training budgets, which are equivalent to 2880, 4800B, and 5760B \layertokens. 

The results are summarized in~\tref{tab:gpt3_1.3B_budgets}. 
The first noticeable result here is that the total amount of training \layertokens directly affects the model's final quality.
Longer training usually leads to better accuracy.
Meanwhile, \OURS can use 2/3 of the training budget to achieve similar evaluation results as the baseline. 
For example, \OURS with training budgets 2880, 4800, and 5760 billion \layertokens can lead to similar accuracy as the baseline with training budgets 4320, 7200, and 8640 billion \layertokens.

\subsection{The interplay between \OURS and Dropout}
\label{sec:the_interplay_between_outs_and_dropout}
Token dropping can be viewed as a special (coarse-grained) case of dropout. 
As such, it may have the potential to work as dropout for certain tasks or further help dropout for certain tasks. 
To investigate this, we apply \OURS with/without standard dropout on both pretraining and finetuning tasks. 

\textbf{\bertlarge Pretraining.} 
Please see~\appref{sec:bert_experimental_setup} for training details. 
For \OURS-3 (\OURS-4), the initial kept token length for all middle layers is 256, and it increases by 16 for every 3.8 (6) billion training tokens such that we eventually save 14.1\% (22.3\%) \layertoken consumption. 

The results are summarized in~\tref{table:bert-ablation-study} with details in~\tref{table:bert-ablation-study-full} and we include the pretraining perplexity to better understand the dropout effect.
Clearly, turning off dropout results in lower evaluation/test perplexity (shown in \tref{table:bert-ablation-study}). 
Meanwhile, the performance of those no-dropout models (baseline*, \OURS-3*, \OURS-4*) on MNLI and QQP show an obvious improvement over their dropout counterparts (baseline, \OURS-3, \OURS-4). 
However, for RACE finetuning, there is no learning for the no-dropout baseline model, which somewhat is surprising but shows the importance of dropout for pretraining. 
In contrast, when turning off the dropout for \OURS, we see a compelling better accuracy on RACE, which  exceeds the standard baseline pretraining by >1\% on RACE-m. 
Thus, \OURS brings not only the efficiency but also the potential regularization effect to \bert pertaining.

\begin{table}[!htb]
\vspace{-0.5cm}
\begin{minipage}[t]{.38\linewidth}
\caption{
Zero-shot  results on \gptb with various  budgets (by \layertoken). 
See~\tref{tab:gpt3_1.3B_budgets_full} for all 19 tasks.
}\centering
\label{tab:gpt3_1.3B_budgets}
\begin{adjustbox}{width=0.8\linewidth}
\centering
\begin{tabular}{lcccccccccccccc}
\toprule
Method & Budget (B) & Ave.  \\
\midrule
\multirow{5}{*}{Baseline} & 2880 & 41.0 \\
                          & 4320 & 41.7 \\
                          & 5760 & 42.5 \\
                          & 7200 & 42.7 \\
                          & 8640 & 43.1 \\
\cdashlinelr{1-8} 
\multirow{3}{*}{\OURS}    & 2880 & 42.1 \\
                          & 4800 & 42.5 \\
                          & 5760 & 43.1 \\
\bottomrule
\end{tabular}
\end{adjustbox}
\end{minipage}\hfill
\begin{minipage}[t]{.6\linewidth}
\caption{
Study the regularization effect of \OURS. \\
We report the average of dev and test for RACE-m and RACE-h. 
* means no dropout. 
Please see~\tref{table:bert-ablation-study-full} for the full result with standard deviation.
}
\label{table:bert-ablation-study}
\begin{adjustbox}{width=0.99\linewidth}
\centering
\begin{tabular}{lcccccccccc}
\toprule
    (Layer-token saving)  & Pretraining PPL val/test  & Downstream Ave. \\\midrule
baseline   (None)              & 5.78/5.80  &  80.14 \\
baseline*   (None)             & 5.45/5.46 & 56.76 \\
\cdashlinelr{1-10}
\OURS-3 (14.1\%)     & 6.37/6.40  & 78.72 \\
\OURS-3* (14.1\%)          & 5.79/5.80 & 81.01  \\
\cdashlinelr{1-10}
\OURS-4  (22.3\% )    & 6.52/6.58 & 78.16 \\
\OURS-4* (22.3\%)  & 6.02/6.04 & 80.49 \\
 \bottomrule
\end{tabular}
\end{adjustbox}
\end{minipage}\hfill
\end{table}

\textbf{\gpthf Finetuning.}
Let us further study the potential regularization effect of \OURS on \gpthf finetuning tasks. 
We train two sets of models, one without dropout and one with dropout (the default rate is 0.1). 
We also apply \OURS with three initial sequence lengths--- 128, 256, and 512---such that they reach the full sequence 1024 at the same epoch (12). 
We present their validation curves in~\fref{fig:regularization}. 

We see that the baseline training without dropout quickly overfits, as shown in the left of~\fref{fig:regularization} (black curve).
And its best validation PPL is worse than the baseline with dropout (block curve in right figure). 
However, for all three cases of \OURS, they achieve similar PPL as the standard baseline with dropout. 
Even with the default dropout in the right of~\fref{fig:regularization}, its validation still faces a non-negligible overfitting issue after the middle of training. 
In contrast, \OURS with \pslg
introduces a potential (extra) regularization effect.
As can be seen, the validation curves of \OURS are flattening and they maintain in a considerably small value regime towards the end of training. 

\textbf{Summary.} 
We do not claim that \OURS can replace dropout. 
In fact, the two can work well (note that both are turned on for the results in GPT pretraining in~\sref{sec:results-pretraining})
with each other as dropout and \OURS focus on different granularity. 
We see that \OURS with dropout achieves lower perplexity for the small finetuning task (shown in the right of~\fref{fig:regularization}). 
Thus, the additional potential regularization effects introduced by \OURS could be complementary to dropout and play a helpful role in those small-dataset tasks with critical overfitting issues.
Also, we see that \OURS without dropout achieves better downstream finetuning accuracy for low-budget \bertlarge pretraining. 
This may indicate \OURS can be a potential regularization method for low-budget pretraining.

%% file: _s6_conclusion.tex
\section{Conclusions}
\label{sec:conclusions}
In this work, we propose a novel random and layerwise token dropping algorithm (\OURS) along with dropping schedules and a new learning rate schedule. 
We demonstrate the efficiency of \OURS on both \gpt and \bert pretraining problems as well as \gpt and \vit finetuning tasks. 
In addition, we probe all ablation studies to verify the effectiveness of each single algorithm component used in \OURS. 
Furthermore, we also show the potential regularization effect introduced by \OURS. 
For the discussion on the limitations and future work, see \appref{sec:future}.

%% file: _s7_appendix.tex
\appendix

%%%%%%%%%%%%%%%%%%%
% Re-count the Figure/Algorithm/Tables after this point. 
%%%%%%%%%%%%%%%%%%%
\counterwithin{figure}{section}
\counterwithin{table}{section}

\section{Other Efficient Training Approaches}
\label{sec:other_efficient_training_approaches}

Mixed-precision training~\citep{micikevicius2017mixed}, different parallelism schemes~\citep{shoeybi2019megatron,huang2019gpipe}, and memory-efficient system design~\citep{rasley2020deepspeed,rajbhandari2020zero} are the most commonly used system-related efficient training methods to train large-scale transformer models. 

Besides those system-level optimizations, researchers and practitioners also investigate various efficient training methods.
\citet{gong2019efficient,li2020shallow} propose layer stacking to speed up \bert training by gradually increasing the number of layers.
\citet{zhang2020accelerating,liu2021faster} extend this idea by adaptively changing the depth of the training. 
Following this, \citet{shen2022staged} further incorporates the growth of the width along with depth for language modelings. 
\citet{rae2021scaling} tile the weights from a small model to a larger one to reduce the training cost for larger models. 
\citet{li2021curriculum} introduce curriculum learning to stabilize the training and get faster convergence of \gpt models. 
\citet{liu2021efficient} propose auxiliary self-supervised task to enable \vit to be effectively trained on small datasets. 

Despite the remarkable success, those methods usually only demonstrate their capability on one specific application and do not show their generalizability across the wide usage of transformer models. 
In contrast, we extensively test \OURS on both pretraining and finetuning.
\section{Compare between baseline, \tokenbypass and \OURS}
\begin{figure}[H]
    \centering
     \includegraphics[width=1.0\linewidth]{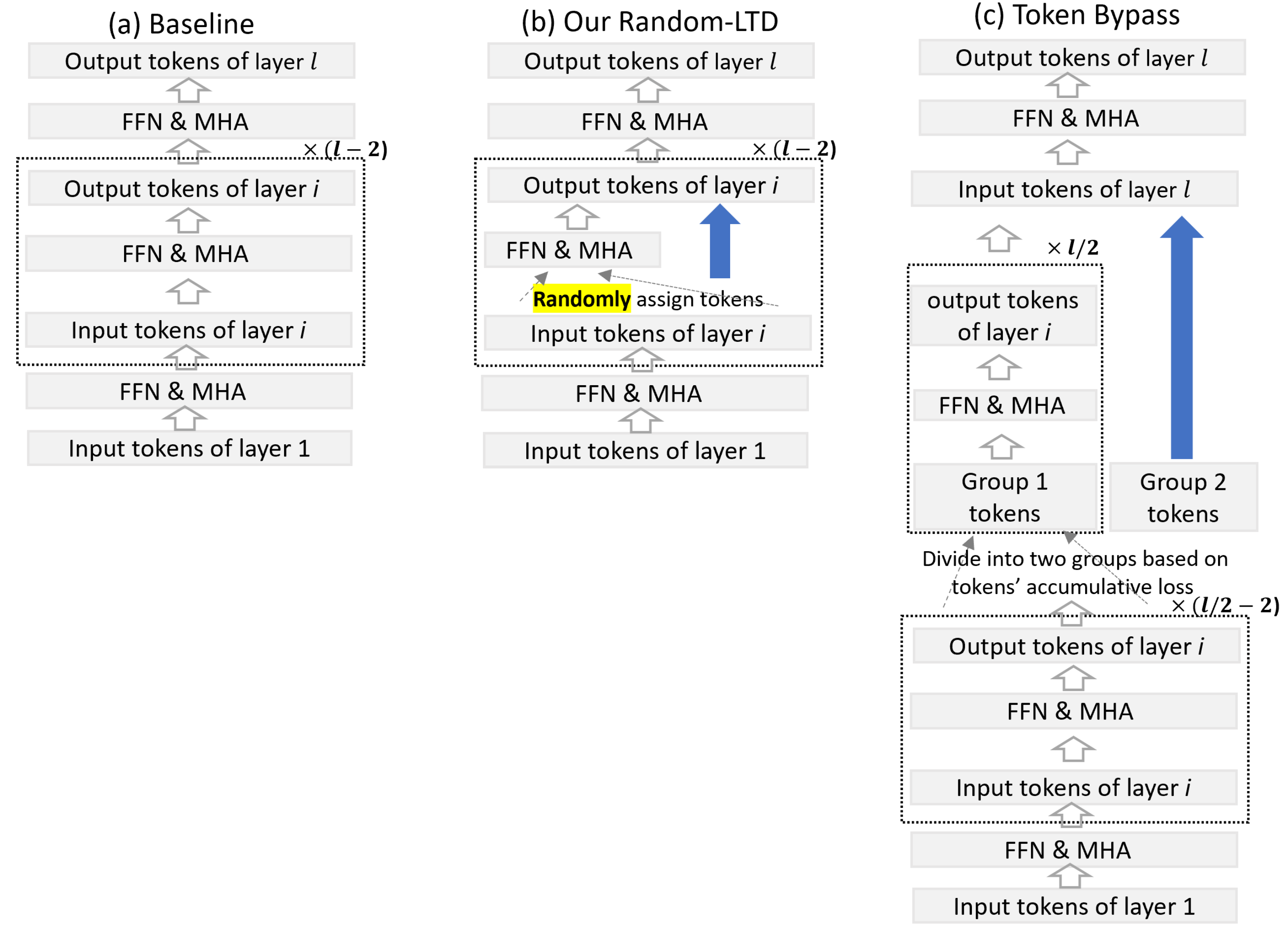}
    \caption{Illustration of the transformer model for  the baseline training (left),  \tokenbypass training (right) and \OURS training (middle). Compared to \tokenbypass, \OURS requires no criterion on the dropped tokens and trains well for all middle layers. The box with dash line is a repeated block. For both (a) and (b), the block is repeated by  $l-2$ times, while for (c), the block is repeated by  $l/2$. In the box, "Output tokens of layer $i$" is the same as "Input tokens of layer $i+1$".   }\label{fig:illustration_of_random_ltd_and_baseline_and_tokenbypass}
\end{figure}

We include the illustration of the comparison between baseline, \tokenbypass, and \OURS in~\fref{fig:illustration_of_random_ltd_and_baseline_and_tokenbypass}.

\section{Formal \layertokenlr Description}
\label{sec:formal_layertokenlr_description}
Formally, recall that the number of drop or kept tokens is $a_{i, t}$ or $b_{i, t}$ for layer $L_i$ at iteration $t$. 
Since we have the same size of $J_i$ and $K_i$ for all layers, we drop the subscript $i$ for simplicity.
At iteration $t$, the total \layertoken consumed by the entire network is $ 2s+(l-2)b_t$ (\OURS is applied to all middle layers). 
Then the total consumed tokens is $2sT+\sum_{t=0}^T(l-2)b_t$. 

Suppose for the iteration-based, the warmup iterations is $T_{\text{warmup}}$, which means the consumed tokens for warmup in all layers is $slT_{\text{warmup}}$. 
Then, the warmup iterations using our \layertokenlr would be $T_{\text{LTwarmup}}$ such that
\begin{equation}
2sT_{\text{LTwarmup}}+\sum_{t=0}^{T_{\text{LTwarmup}}}(l-2)b_t=slT_{\text{warmup}}.
\end{equation}
Similarly, for an iteration-based learning rate schedule, the decay learning schedule (e.g., linear decay or cosine decay schedules) is based on the length of $T-T_{\text{warmup}}$. 
For our \layertokenlr schedule, the corresponding length is
\begin{equation}
2sT + \sum_{t=0}^T(l-2)b_t - 2sT_{\text{LTwarmup}} - \sum_{t=0}^{T_{\text{LTwarmup}}}(l-2)b_t.
\end{equation}
Note that this proposed schedule would reduce to standard learning rate if we keep $b_t=s$ for all layers (without \OURS). 
Please see~\sref{sec:lr_schedule_effect} for the effectiveness of our \layertokenlr as compared to standard learning rate schedule and see~\fref{fig:token-lr-gpt} for the learning rate schedule illustration.

\section{Detailed Experimental Setup}

\subsection{ Experimental Setup for \gpt Pretraining}
\label{sec:gpt_experimental_setup}
We use DeepSpeed~\citep{rasley2020deepspeed} and Megatron-DeepSpeed~\citep{Megatron-DeepSpeed} repository to train \gpt models with 350 million parameters (\gptm) and 1.3 billion parameters (\gptb).
The pretraining data are from PILE dataset~\cite{gao2020pile} and the total training tokens are 300 billion without extra explanation. 
For \OURS, the initial dropped token length for all middle layers is 1920 (i.e., 128 tokens are kept for compute), and it decreases by 16 for every 1.75 billion training tokens. 
That is to say, after 210B training tokens, \OURS degrades to standard training procedure with full sequence length. 
Theoretically, this can save 1/3 of the layer-token training budget. 
All models are trained with 64 A100-40G GPUs.

We evaluate our results on 19 zero-shot evaluation tasks, including 19 accuracy evaluation tasks (i.e., HellaSwag~\citep{zellers2019hellaswag}, LAMBADA~\citep{paperno2016lambada}, TriviaQA~\citep{joshi2017triviaqa}, WebQS~\citep{berant-etal-2013-semantic}, Winogrande~\citep{sakaguchi2020winogrande}, PIQA~\citep{tata2003piqa}, ARC (Challenge/Easy)~\citep{boratko2018systematic}, ANLI (R1/R2/R3)~\citep{williams2020anlizing}, OpenBookQA~\citep{mihaylov2018can}, RACE-h~\citep{lai2017race}, BoolQ~\citep{clark2019boolq}, Copa~\citep{afshar2018copa}, RTE~\citep{dagan2013recognizing}, WSC~\citep{levesque2012winograd}, MultiRC~\citep{yadav2019quick}, and ReCoRD~\citep{zhang2018record}). 

\subsection{Experimental Setup for \bert Pretraining}
\label{sec:bert_experimental_setup}
Similar to \gpt pretraining, we use the Megatron-DeepSpeed repository to train our \bertlarge models with 336 million parameters (24 layers) with sequence length 512. The pretraining recipe follows \citep{shoeybi2019megatron}, but the pretraining data is public, the same as \gpt pretraining.  

For the  main results in \sref{subsec:bert-pretraining}, the total training tokens is $512\times1024\times2\times10^6$.
Here 1024 is the global batch size and 512 is the sequence length. We pretrain with 2 million iterations.
We trained on 64 A100-40g GPUs (16 batch/GPU). 
For \OURS-2 (\OURS-1), the initial kept token length for all middle layers is 128 (200) for compute, and it increases by 16 for every 38 (48) billion training tokens such that we eventually save 31.1\% (26.23\%) layer-tokens.  
We take the last checkpoint to perform downstream tasks on the popular stable benchmarks, including MNLI~\citep{williams2017broad}, QQP~\citep{iyer2017first}, and RACE (middle and high difficulty)~\citep{lai2017race}, of which the fine-tuning configurations (same as~\citep{shoeybi2019megatron}) respectively are 10, 12 and 3 epochs with batch-size 128, 128 and 32 and learning rate 1e-5, 5e-5 and 2e-5. 
We report the median (best) of five repeated runs (random seeds 1234-1238) in~\tref{table:bert-main}. 
Note that we apply the standard training  (no \OURS) in the downstream tasks in order to have fair comparisons.

For the ablation study of \bert pretraining in \sref{sec:no_need_special_token_treatment} and~\ref{sec:the_interplay_between_outs_and_dropout},  we shorten the pretraining iteration to 0.2 million (due to limited resource) and set the maximum learning rate to be 4e-4. We start with the initial sequence 256 and increase it by 16 for every 6 billion training tokens such that we eventually save 22.3\% layer-tokens. To compensate for the short-time pretraining, we make the standard fine-training epoch much longer (30 epochs) for MNLI, QQP, and RACE and their batch-size  (learning rate) are 128 (5e-5), 128 (5e-5), and 64 (2e-5) respectively. 

\subsection{Experimental Setup for \vit Fine-tuning}\label{subsec:fine-tune-vit}

We apply \OURS to the vision transformer (\vit)~\citep{dosovitskiy2021an} on fine-tuning tasks in order to demonstrate the broader applications of our method across different domains. 
We use the pretrained models published in~\citep{rw2019timm} and focus mainly on the two small image recognition benchmarks--- CIFAR10 and CIFAR100~\citep{krizhevsky2009learning}, 
and one large-scale dataset---ImageNet~\citep{deng2009imagenet}. 
For ImageNet (CIFAR10/100), we use the 12-layer (24-layer) pretrained \vit with an input resolution $224\times 224$ in which each patch of size $16\times 16$ such that the sequence length becomes $196+1$ (the extra token is for position). 
ImageNet (CIFAR10/100) is trained on an 8-GPU (1-GPU) A100-40G machine such that the batch size is 32 (128) images per GPU. 
The training budget for all three datasets is 14 epochs and a small constant learning rate is used based on grid search.
Particularly, the best learning rate for ImageNet is 5e-5 and CIFAR10/CIFAR100 is 1e-4. For ImageNet (CIFAR), the sequence length is started with 66 (32) and linearly reaches to the 197 full sequence length at 80\% of the total training iterations such that  22.3\%(30.9\%) layer-token saving is achieved.

\subsection{Experimental Setup for GPT Finetuning}\label{subsec:fine-tune-gpt}

For the language fine-tuning tasks, we directly take the existing  pretrained GPT model (350M, 24-layer) published in HuggingFace \citep{wolf2019huggingface} and fine-tune on the three dataset: Penn Treebank (PTB) \citep{marcus-etal-1993-building}, WikiText-2 and WikiText-103 \citep{merity2017pointer}. These fine-tuning tasks are trained on a single V100 GPU with a batch size of 32 and a constant learning rate 5e-5. We trained for 15 epochs for PTB, 4 epochs for WikiText-103  and 10 epcoh WikiText-2. As for \OURS in \tref{table:main-layers}, the sequence length is started with 128 (64) sequence  with a linear increase to the 1024 full sequence length at 80\% (70\%) of the total training iterations for PTB (WikiText-103/-2).

\section{Standard deviation for \bertlarge  Downstream tasks}
We show the standard deviation for \bertlarge downstream tasks finetuning in~\tref{table:bert-std}.

\begin{table}[H]
\caption {The comparison between baseline and \OURS on finetuning for \bertlarge. Complementary to \tref{table:bert-main}, we report the mean and one standard deviation over five independent runs with the same hyperparameters.
}\label{table:bert-std}
\begin{adjustbox}{width=0.99\linewidth}
\centering
\begin{tabular}{lccccccccccc}
\toprule
  Method     & MNLI-m/-mm             & QQP               & RACE-m (dev) &   RACE-m (test) & RACE-h (dev)  & RACE-h (test)  \\\midrule
Baseline   & 89.09±0.05/89.52±0.21& 92.29±0.12& 84.58±0.25& 82.88±0.57& 80.54±0.29& 79.01±0.45\\
\cdashlinelr{1-9}
\OURS-1    & 89.73±0.09/89.93±0.15& 92.09±0.08& 85.67±0.32& 84.51±0.58& 82.78±0.28& 81.47±0.43\\
 \OURS-2   &89.36±0.1/89.74±0.12& 91.96±0.11& 86.14±0.29& 84.96±0.27& 82.08±0.19& 80.74±0.16\\
\bottomrule
\end{tabular}
\end{adjustbox}
\end{table}

\section{\bertlarge pretraining with/without \layertokenlr}
\label{sec:bert-lr}
In~\sref{sec:lr_schedule_effect}, we have seen the benefits of using \layertokenlr over the standard learning rate for GPT pretraining. 
Here we present the additional results for \bertlarge pretraining. 
The training budget is 0.2 million iterations, and the training details are given in~\appref{sec:bert_experimental_setup}. 
The results are presented in~\tref{table:lr-bert}. which further confirms the dominant benefit of \layertokenlr.
\begin{table}[H]
\caption{Results of \bertlarge pretraining with 0.2 million iterations for \OURS with  14.1\% layer-token saving. Random-LTD-3* in \tref{table:bert-ablation-study} is the same as the last row (\layertokenlr). }\label{table:lr-bert}
\begin{adjustbox}{width=0.99\linewidth}
\centering
\begin{tabular}{l|cc|cccccccc}
\toprule
  Learning rate     & \multicolumn{2}{c|}{ Pretraining results}  & \multicolumn{6}{c}{ DownStream finetuning results} \\
 method     & ppl(val)  & ppl(test)  & MNLI-m/-mm & QQP    & RACE-m (dev) &   RACE-m (test) & RACE-h (dev)  & RACE-h (test)  \\\midrule

Standard LR         &  5.90  & 5.92 & 86.46/86.53& 91.82& 72.09& 73.79& 73.50 & 71.04\\
\layertokenlr       & 5.79   & 5.80 & 87.14/87.29& 91.95& 78.05& 77.84& 73.47& 71.30  \\
 \bottomrule
\end{tabular}
\end{adjustbox}
\end{table}

\section{Limitations and Future Work}\label{sec:future}
We believe it is critical for every work to clearly state its limitations, especially in this area. 
An important limitation of this work is that we keep the dropped token ratio for all intermediate layers the same. 
We design it in such a way as to reduce the parameter tuning effort.
However, each layer may have its sensitivity (see \fref{fig:sensitivity}). 
Therefore, an automated dropped ratio could help here.
Another limitation is that our \pslg is based on a linear increasing manner. 
This might not be optimal, and a self-adaptive schedule could further improve the efficiency and convergence behavior. Finally, in this work, we found out \OURS can have the potential regularization effect as the standard dropout. 
However, to improve its generalizability, there are still a lot of experiments to be done, which is out of the scope of this paper. 
We leave this as future work.

\section{Full results used in main text}
We include the full results used in main text in this section.
\begin{table}[H]
\centering
 \caption{Finetuning result of \vit on ImageNet and CIFAR.
 This is the full result of~\tref{table:vit-main}.} 
 \label{table:vit-main-full}
 \begin{adjustbox}{width=0.99\linewidth}
\begin{tabular}{l|ccc|ccc}
\toprule
           & \multicolumn{3}{c|}{ImageNet datasets on 12-layer ViT} & \multicolumn{3}{c}{CIFAR datasets on 24-layer ViT}            \\
           & \layertoken Saving & Top-1       & Top-5        & \layertoken Saving & Top-1 (CIFAR100) & Top-1 (CIFAR10) \\\midrule
baseline   & N/A            & 84.65±0.04 & 97.41±0.02 & N/A          & 93.93±0.30    & 99.32±0.05     \\
random-LTD & 22.3\%      & 84.70±0.04  & 97.48±0.02 & 30.9\%      & 94.02±0.40    & 99.30±0.03 \\\bottomrule
\end{tabular}
\end{adjustbox}
\end{table}

\begin{table}[H]
\caption{Ablation study of special token treatment for \bert pretraining with 22.2\% \layertoken saving.
This is the full result of~\tref{table:bert-ablation-study-token}.
}\label{table:bert-ablation-study-token-full}
\centering
\begin{adjustbox}{width=0.69\linewidth}
\begin{tabular}{c|cc|cccccc}
\toprule
     Keep    &  \multicolumn{2}{c|}{ Pretraining results}  & \multicolumn{6}{|c}{ DownStream finetuning results} \\
   Special Tokens  & ppl(val)  & ppl(test)  & MNLI-m/-mm & QQP     \\\midrule

yes      & 6.024   & 6.049& 86.70/86.97& 91.83\\
 no          & 6.018  & 6.040& 86.66/86.92& 91.97\\
 \bottomrule
\end{tabular}
\end{adjustbox}
\end{table}

\begin{table}[tb]
 \caption{Comparison of applying \OURS to different layers on \gpthf finetuning and \vit finetuning.
 This is the full result of~\tref{table:main-layers}.
 } 
 \label{table:main-layers-full}
\centering
 \begin{adjustbox}{width=0.89\linewidth}
\begin{tabular}{cccccc}
\toprule
   &  &\multicolumn{4}{c}{Random-LTD applied to all layers except for the following }        \\
  Method  & dataset & None & First &  Last & First and Last                     \\
\midrule
\multirow{3}{*}{Perplexity} & PTB  & 16.00±0.02 & 16.01±0.03  & 16.09±0.02 & 15.92±0.02    \\
 &WikiText-2  &  17.06±0.02 & 17.01±0.02 & 17.01±0.02 & 16.94±0.01    \\
 &WikiText-103  &13.27±0.01  &13.03±0.03 & 13.23±0.04 &12.99±0.01     \\
 \midrule
\multirow{1}{*}{Accuracy}  & ImageNet-Top1  & 84.47±0.08 & 84.51±0.08 & 84.65±0.04  & 84.70±0.04     \\
 \bottomrule
\end{tabular}
\end{adjustbox}
\end{table}

\begin{table}[t]
\centering
\caption{Compare between \pslg and constant token dropping schedules.  This is the full result of~\tref{table:seq-schedules}.}\label{table:seq-schedules-full}
 \begin{adjustbox}{width=0.99\linewidth}
\begin{tabular}{l|cccc|ccc}
\toprule
datasets       & \multicolumn{4}{c|}{CIFAR10 (Metric: Top-1 accuracy) }                         & \multicolumn{3}{c}{PTB  (Metric: perplexity)} \\
Token-drop schedules &  constant & constant &  constant   & \pslg   &   constant & constant  & \pslg\\
\layertoken saving   & 16.5\%      & 23.6\%   & 30.8\%        & 32.3\%  & 23.0\%   & 32.1\%      & 33.7\%       \\\midrule
Performance     & 99.33±0.01      &99.28±0.01  &99.26±0.08  &99.32±0.03 &     18.27±0.08   &     20.76±0.06        &   15.92±0.02  \\
\bottomrule
\end{tabular}
\end{adjustbox}
\end{table}

\begin{table}[t]
\caption{
Study the regularization effect of \OURS. 
We report the average of dev and test for RACE-m and RARCE-h. 
* means no dropout. 
}
\label{table:bert-ablation-study-full}
\begin{adjustbox}{width=0.99\linewidth}
\centering
\begin{tabular}{l|cc|cccccccc}
\toprule
     Method   & \multicolumn{2}{c|}{ Pretraining results}  & \multicolumn{6}{c}{ DownStream finetuning results} \\
    (Layer-token saving)  & ppl(val)  & ppl(test)  & MNLI-m/-mm & QQP    & RACE-m (dev) &   RACE-m (test) & RACE-h (dev)  & RACE-h (test)  \\\midrule
baseline   (None)              & 5.78     & 5.80  &  86.44/86.51& 92.11& 75.85& 75.50& 73.78& 70.78 \\
baseline*   (None)             & 5.45  & 5.46 & 86.4/86.93& 92.07& 32.67& 32.46& 34.52& 32.26  \\\cdashlinelr{1-10}
\OURS-3 (14.1\%)     & 6.37   & 6.40  & 85.91/85.97& 91.84& 74.72& 73.22& 70.95& 68.43  \\
\OURS-3* (14.1\%)          & 5.79   & 5.80 & 87.14/87.29& 91.95& 78.05& 77.84& 73.47& 71.30  \\\cdashlinelr{1-10}
\OURS-4  (22.3\% )    & 6.52  & 6.58 & 85.69/85.64& 91.74& 73.65& 71.88& 70.49& 68.00\\
\OURS-4* (22.3\%)  & 6.02   & 6.04 & 86.66/86.92& 91.97& 76.78& 76.56& 73.38& 71.15 \\
 \bottomrule
\end{tabular}
\end{adjustbox}
\end{table}

\section{Full Zero-shot Evaluation of \gpt-style Models}

We include all zero-shot evaluation results  for all \gpt models in \tref{tab:gpt3_main_result_full} and \ref{tab:gpt3_1.3B_budgets_full}.

\begin{table}[t]
\caption{
Zero-shot evaluation results of baseline and \OURS on \gptm and \gptb. 
}\centering
\label{tab:gpt3_main_result_full}
\begin{adjustbox}{width=0.99\linewidth}
\centering
\begin{tabular}{lcccccccccccccccccccccccccccccccccccc}
\toprule
\multirow{2}{*}{Tasks}  & \multicolumn{2}{c}{Baseline}  & & \multicolumn{2}{c}{\OURS}   \\
\cline{2-3}\cline{5-6}
& \gptm & \gptb & & \gptm & \gptb  \\ 
HellaSwag         & 39.3 & 52.1 & & 40.2 & 51.8\\
LAMBADA           & 52.3 & 61.2 & & 52.3 & 62.2\\
TriviaQA          &  3.6 &  6.3 & & 3.14 & 6.05\\
WebQs             & 1.82 & 2.21 & & 1.58 & 1.92\\
Winogrande        & 53.3 & 55.7 & & 50.9 & 58.0\\
PIQA              & 67.0 & 71.1 & & 67.2 & 71.0\\
ARC (Challenge)   & 25.2 & 29.4 & & 24.9 & 28.2\\
ARC (Easy)        & 46.0 & 53.4 & & 44.4 & 53.2\\
ANLI R1           & 32.7 & 32.9 & & 33.1 & 33.5\\
ANLI R2           & 32.6 & 33.7 & & 33.0 & 32.4\\
ANLI R3           & 33.8 & 35.1 & & 33.7 & 34.6\\
OpenBookQA        & 28.8 & 33.4 & & 29.6 & 32.6\\
RACE-h            & 29.3 & 33.4 & & 31.8 & 34.2\\
BoolQ             & 55.6 & 56.4 & & 58.4 & 62.7\\
Copa              & 67.0 & 71.0 & & 68.0 & 72.0\\
RTE               & 51.6 & 56.7 & & 53.1 & 52.7\\
WSC               & 36.5 & 43.3 & & 36.5 & 36.5\\
MultiRC           & 0.84 & 0.84 & & 0.84 & 0.84\\
ReCoRD            & 76.3 & 82.3 & & 76.4 & 82.9\\
\midrule
Average Acc       & 38.6 & 42.7 & & 38.9 & 42.5\\
\bottomrule
\end{tabular}
\end{adjustbox}
\end{table}

\begin{table}[t]
\caption{
Zero-shot evaluation results of baseline and \OURS on \gptb with various training budgets.
Here, "Budget" in the first column means the training layer-token as the final budget.
}\centering
\label{tab:gpt3_1.3B_budgets_full}
\begin{adjustbox}{width=0.99\linewidth}
\centering
\begin{tabular}{lcccccccccccccccccccccccccccccccccccc}
\toprule
\multirow{2}{*}{Tasks/Budgets (B)}  & \multicolumn{5}{c}{Baseline}  & & \multicolumn{2}{c}{\OURS}   \\
\cline{2-6}\cline{8-10}
& 2880 & 4320 & 5760 & 7200 & 8640 & & 2880 & 4800 & 5760  \\ 
HellaSwag         &46.4	&49.2	&51.9	&52.1	&52.9	&	&49.1	&51.8	&53.4\\
LAMBADA           &57.1	&59.0	&60.2	&61.2	&61.5	&	&60.2	&62.2	&62.8\\
TriviaQA          &5.36	&5.58	&7.73	&6.3 	&7.2 	&	&4.79	&6.05	&7.46\\
WebQs             &2.31	&2.02	&1.33	&2.21	&2.26	&	&1.97	&1.92	&1.72\\
Winogrande        &54.7	&55.1	&57.5	&55.7	&56.6	&	&56.4	&58.0	&60.1\\
PIQA              &70.1	&70.1	&71.1	&71.1	&72.3	&	&70.5	&71.0	&70.7\\
ARC (Challenge)   &26.0	&26.6	&27.8	&29.4	&28.8	&	&26.2	&28.2	&28.9\\
ARC (Easy)        &51.1	&52.0	&52.5	&53.4	&54.6	&	&51.6	&53.2	&53.2\\
ANLI R1           &33.0	&31.3	&32.7	&32.9	&32.1	&	&33.4	&33.5	&33.2\\
ANLI R2           &32.6	&32.2	&35.7	&33.7	&33.4	&	&32.7	&32.4	&33.0\\
ANLI R3           &35.2	&33.8	&36.9	&35.1	&35.8	&	&34.8	&34.6	&32.9\\
OpenBookQA        &31.2	&31.2	&33.0	&33.4	&33.6	&	&33.6	&32.6	&34.2\\
RACE-h            &32.8	&34.4	&34.0	&33.4	&36.6	&	&34.3	&34.2	&35.7\\
BoolQ             &62.0	&58.9	&61.5	&56.4	&59.9	&	&61.7	&62.7	&62.8\\
Copa              &71.0	&72.0	&70.0	&71.0	&74.0	&	&73.0	&72.0	&71.0\\
RTE               &52.0	&58.1	&54.2	&56.7	&56.3	&	&56.7	&52.7	&56.3\\
WSC               &36.5	&36.5	&36.5	&43.3	&36.5	&	&36.5	&36.5	&37.5\\
MultiRC           &0.94	&2.62	&0.94	&0.84	&1.26	&	&0.84	&0.84	&0.84\\
ReCoRD            &79.5	&81.2	&82.2	&82.3	&83.2	&	&81.6	&82.9	&83.4\\
\midrule
Average Acc       &41.0	&41.7	&42.5	&42.7	&43.1	&	&42.1	&42.5	&43.1\\
\bottomrule
\end{tabular}
\end{adjustbox}
\end{table}